\documentclass[sigconf]{acmart}

\usepackage{todonotes}
\usepackage{multirow}
\usepackage{subcaption}
\usepackage{colortbl}

\acmConference[FDG '22]{International Conference on the Foundations of Digital Games}{September 05--08,
  2022}{Athens, Greece}
\acmPrice{15.00}
\acmISBN{978-1-4503-XXXX-X/18/06}

\begin{document}

\title{Persona-driven Dominant/Submissive Map (PDSM)\\Generation for Tutorials}

\author{Michael Cerny Green}
\email{mike.green@nyu.edu}
\orcid{0000-0003-3366-8165}
\affiliation{%
  \institution{New York University}
  \city{New York}
  \state{New York}
  \country{USA}
}

\author{Ahmed Khalifa}
\email{ahmed.khalifa@um.edu.mt}
\affiliation{%
  \institution{University of Malta}
  \city{Msida}
  \country{Malta}
}

\author{M Charity}
\email{mlc761@nyu.edu}
\affiliation{%
  \institution{New York University}
  \city{New York}
  \state{New York}
  \country{USA}
}
\author{Julian Togelius}
\email{julian@togelius.com}
\affiliation{%
  \institution{New York University}
  \city{New York}
  \state{New York}
  \country{USA}
}
\renewcommand{\shortauthors}{Green, et al.}

\begin{abstract}
In this paper, we present a method for automated persona-driven video game tutorial level generation. Tutorial levels are scenarios in which the player can explore and discover different rules and game mechanics. Procedural personas can guide generators to create content which encourages or discourages certain playstyle behaviors. In this system, we use procedural personas to calculate the behavioral characteristics of levels which are evolved using the quality-diversity algorithm known as Constrained MAP-Elites. An evolved map's quality is determined by its simplicity: the simpler it is, the better it is. Within this work, we show that the generated maps can strongly encourage or discourage different persona-like behaviors and range from simple solutions to complex puzzle-levels, making them perfect candidates for a tutorial generative system.
\end{abstract}



\maketitle


\section{Introduction}

Most games include some form of tutorial that teaches you how to play the game. In particular, games often feature levels that teach the player various aspects of the game. It is very common for such tutorial content to occur at the beginning of the game, but also whenever new mechanics, NPC types or similar concepts are introduced. For example, when you find the grapple hook in one of the Zelda games, you can be certain to find a few puzzles nearby that requires you to use it so that you can learn by doing. 

Many games feature some form of level generation, and procedural content generation for games is also an active research field~\cite{shaker2016procedural}. In this context, the question of how to generate tutorials has attracted some attention recently~\cite{green2018generating,green2018atdelfi,green2020mario,aytemiz2018talin}. The reasons for wanting to generate tutorials are multiple: one might want to part-automate the tutorial creation process to free up game designers and developers to work on other topics; one might want to enable automatically generated games to have tutorials; and one might want to create tutorial-generating systems in order to better understand tutorials in general. But there is yet another reason for creating systems that generate tutorials: the ability to generate tutorials for specific players of player types, to fit their playstyle and/or capabilities. Such systems could create tutorials that are more engaging and effective for particular players, instead a one-size-fits-all approach that becomes more or less necessary with hand-crafted tutorials. In some of the research that this paper builds on, tutorial levels were generated that taught specific mechanics, in the sense that a player needed to use (or not use) specific mechanics in order to finish the level~\cite{green2018generating,khalifa2018talakat,charity2020mech}.

Generating tutorials to fit playstyles can be seen as a form of experience-driven procedural content generation (EDPCG)~\cite{yannakakis2011experience}. In the EDPCG paradigm, models of user experience (and potentially also player behavior) are used to guide content generation algorithms. In practice this can take the shape of fitness functions for evolutionary algorithms, rewards for reinforcement learning algorithms, or constraints for constraint satisfaction algorithms. For example, one could learn a predictor of engagement that uses playstyle characteristics and level characteristics as input; by keeping the playstyle constant, one could use evolutionary computation to search for levels that maximize engagement for a particular player~\cite{shaker2010towards,shaker2012crowdsourcing}.

Following this paradigm, there are several ways in which one could create a tutorial level that is tailored to a specific playstyle. One way is to create a level that mandates the use of some mechanic (or in general, overcomes a specific type of challenge) and that is winnable using the currently used playstyle. Another type of tutorial level is one that requires the player adopt a different playstyle to win. In the latter case, the playstyle may be something the game wishes to teach in order to enable the player to enjoy more of the game. Also, the game can try to teach when to avoid a certain playstyle to help the player understand the strategic depth of the game.

In this paper, we investigate methods for generating tutorial levels tailored to playstyles in MiniDungeons 2~\cite{holmgaard2015minidungeons}. For this, we use the concept of procedural personas. A procedural persona is a generative model (i.e., a game-playing agent) of an archetypical playstyle~\cite{holmgaard2014evolving,holmgaard2018automated}. The three procedural personas used in this study -- the runner, the monster killer, and the treasure collector -- exemplify sharply different strategies of playing a game. We then use a quality-diversity algorithm, Constrained MAP-Elites, to find large sets of different levels. The levels systematically vary depending on which personas can perform well on them. Thus, the system can generate levels that can be played equally well by multiple playstyles, that require the use of a particular playstyle, or that require the avoidance of a particular playstyle.

\section{Background}
In this section, we discuss the usage of quality diversity algorithm for Procedural level generation and usage of player personas as content evaluators not only in MiniDungeons 2 (the game used in this research) but also in other games. We review recent research in experience-driven procedural content generation, touch upon different types of video game tutorial patterns, and describe the MiniDungeons 2 framework in detail.

\subsection{Experience-driven PCG}
Procedural Content Generation (PCG)~\cite{hendrikx2013procedural,shaker2016procedural} is the process of using computer algorithms to produce content. PCG techniques have been utilized in games as far back as the level generation systems in Beneath the Apple Manor (Don Worth, 1978) and Rogue (Glenn Wichman, 1980), and continue to be used to generate maps in Spelunky (Derek Yu, 2008), terrain in Minecraft (Mojang, 2011), and worlds in Starbound (Chucklefish, 2016) and No Man's Sky (Hello Games, 2016).
Most PCG methods generate content for the general audience of the game in question. Experience-driven Procedural Content Generation~\cite{yannakakis2011experience} (EDPCG) can content designed for a specific user experiences. EDPCG may use any of the methods specified in previous sections for a wide variety of applications such as generating levels~\cite{shu2021experience}, music~\cite{plans2012experience}, and therapeutic experiences~\cite{mahmoudi2021arachnophobia}. Designing for a specific user experience can come in many forms, such as level difficulty (as measured by the player's skill)~\cite{huber2021dynamic}, aesthetics~\cite{liapis2011optimizing}, or letting users pick for themselves~\cite{liapis2013adaptive}.

\subsection{Quality-Diversity PCG}\label{background-qdpcg}
Quality diversity algorithms (QD) have been seeing increasing use in PCG, especially in level generation~\cite{gravina2019procedural}. The main reason behind that is the designer doesn't need to code complex fitness function any more. Instead, all the complexity can be moved towards the diversity component of the algorithm while only focusing on objective qualities like having a playable playable. When the algorithm is done, it ends with a huge amount of different levels that not diverse and different but also maintain certain quality. Several methods have been used to generate varies types of contents. For example, using novelty search with local competition to generate a variety of The Sims houses such that a sim's agent could stay alive~\cite{charity2020say}. Garvina et al.~\cite{gravina2016constrained} used constrained surprise search to generate balanced weapons in the game of Unreal Tournament III such that the generated weapons have different behavior than previously discovered ones.

One of the most well-known QD algorithms is MAP-Elites~\cite{mouret2015illuminating}, an algorithm that uses a multi-dimension grid instead of population to store evolved solutions. A set of distinct behavior characteristics determines where in the map a solution resides, while a solution's fitness determines if it will be saved or thrown away. MAP-Elites has been used in a variety of PCG projects such as generating Hearthstone Decks~\cite{fontaine2019mapping,zhang2021deep}, 2D and 3D objects~\cite{lehman2016creative,nguyen2015innovation}, and platformer levels~\cite{warriar2019playmapper}. Constrained MAP-Elites (CME) is a hybrid algorithm that combines MAP-Elites with the FI2-Pop algorithm~\cite{kimbrough2008feasible} that was introduced for bullet hell level generation~\cite{khalifa2018talakat}. CME has been used for other projects such as mixed-initiative generation of dungeon crawlers~\cite{alvarez2019empowering} and tutorial generation of levels that teach certain game mechanics~\cite{khalifa2019intentional,charity2020mech}.




\subsection{Procedural Personas}\label{sec:background-personas}
Bartle~\cite{bartle1996hearts} originally proposed a taxonomy of players and their personas based on how they interacted with the environment and other players ranging from killers, socializers, achievers, and explorers. Typically, designers try to design levels to cater to a set of personas or to design levels with multiple paths that work with different personas.
Unsupervised learning methods have been used to infer play personas using playtraces from Tomb Raider: Underworld~\cite{drachen2009player} and Starcraft~II~\cite{avontuur2013player}. Meanwhile supervised learning methods have been used to infer play personas~\cite{green2022predicting} using either playtraces or mechanic activation counts from MiniDungeons 2.

Tychsen and Canossa~\cite{tychsen2008defining} introduced player personas through game metrics and triggered mechanics. These metrics could be used to recreate a specific player but may not give as much insight as to what a type of player might do on a general scale (i.e. how would the same player react in a game level they have never played before.) With automated and artificial personas, these agents can be inserted into the game to predict how a player may approach a game level, saving human and computation resources and speeding up the design process. Procedural agents developed via evolution~\cite{holmgaard2014evolving,holmgaard2018automated} and reinforcement learning~\cite{holmgard2014generative} have shown to be accurate in emulating player behaviors in a game setting. These artificial personas can be used for automated playtesting and encapsulate a variety of behaviors which lead to a wider diversity of levels designed for players with different play personas.

\subsection{Video Games Tutorials}
Previous work on automated tutorial generation has highlighted common tutorial patterns found across games. \emph{Instruction-based tutorials} explain how to play the game by providing the player with a group of instructions to follow, similar to what is seen in boardgames. For example: Strategy games, such as \emph{Starcraft} (Blizzard, 1998), teach the player by taking them step by step towards understanding different aspects of the game.
\emph{Demonstration-based tutorials} show the player an example of what will happen if they do a specific action. For example: Megaman X uses a Non Playable Character (NPC) to teach the player about the charging skill~\cite{egoraptor2011megaman}.
\emph{Experienced-based tutorials} provide the player freedom to explore and experiment using a carefully designed level or level-sequence. For example: in \emph{Super Mario Bros} (Nintendo, 1985), the world 1-1 is designed to introduce players to different game elements, such as goombas and mushrooms, in a way that the player can not miss~\cite{credits2014supermario}. In this work, we focus on the automated generation of experience-based tutorials via level generation. We build upon previous work in this space by using personas rather than mechanics~\cite{charity2020mech,khalifa2019intentional,green2020mario} or skill~\cite{aytemiz2018talin} to guide the output of the generator.

\subsection{MiniDungeons 2}\label{sec:background-md2}

\begin{figure}
    \centering
    \includegraphics[height=40ex]{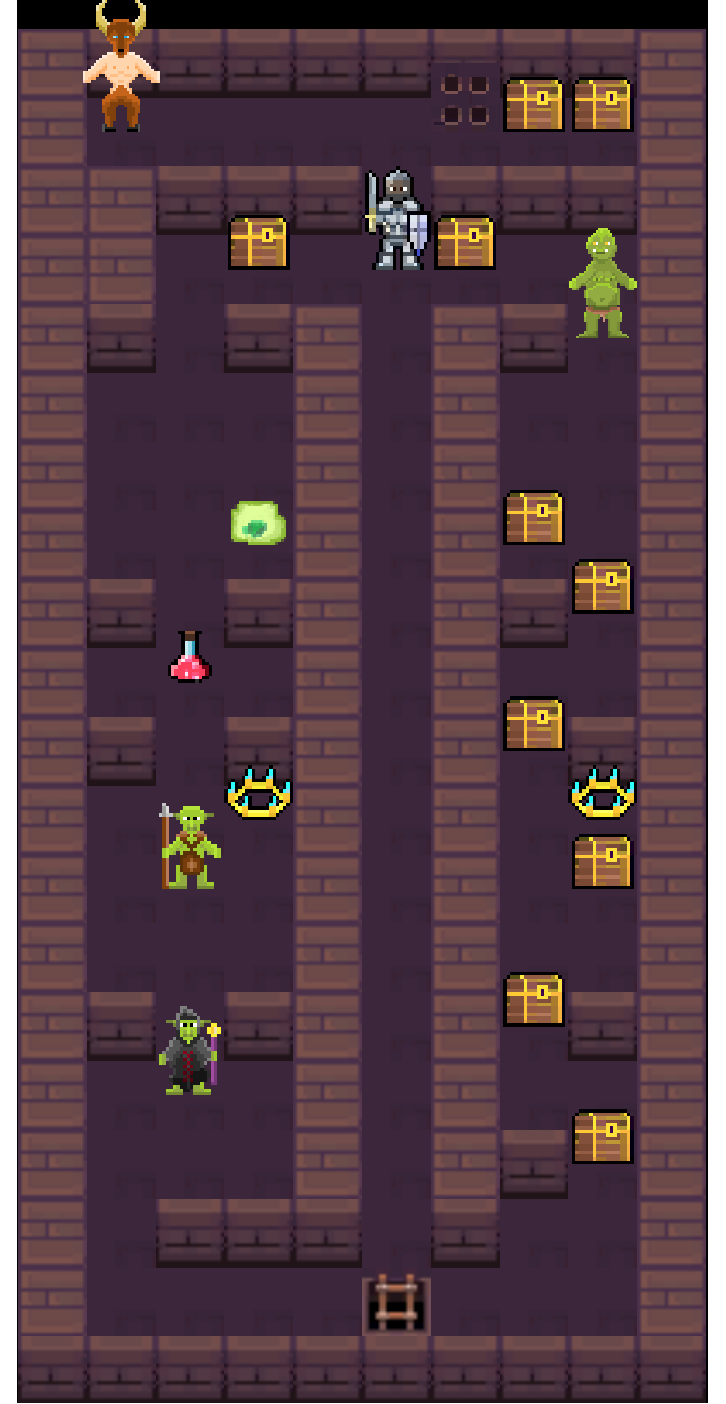}
    \includegraphics[height=40ex]{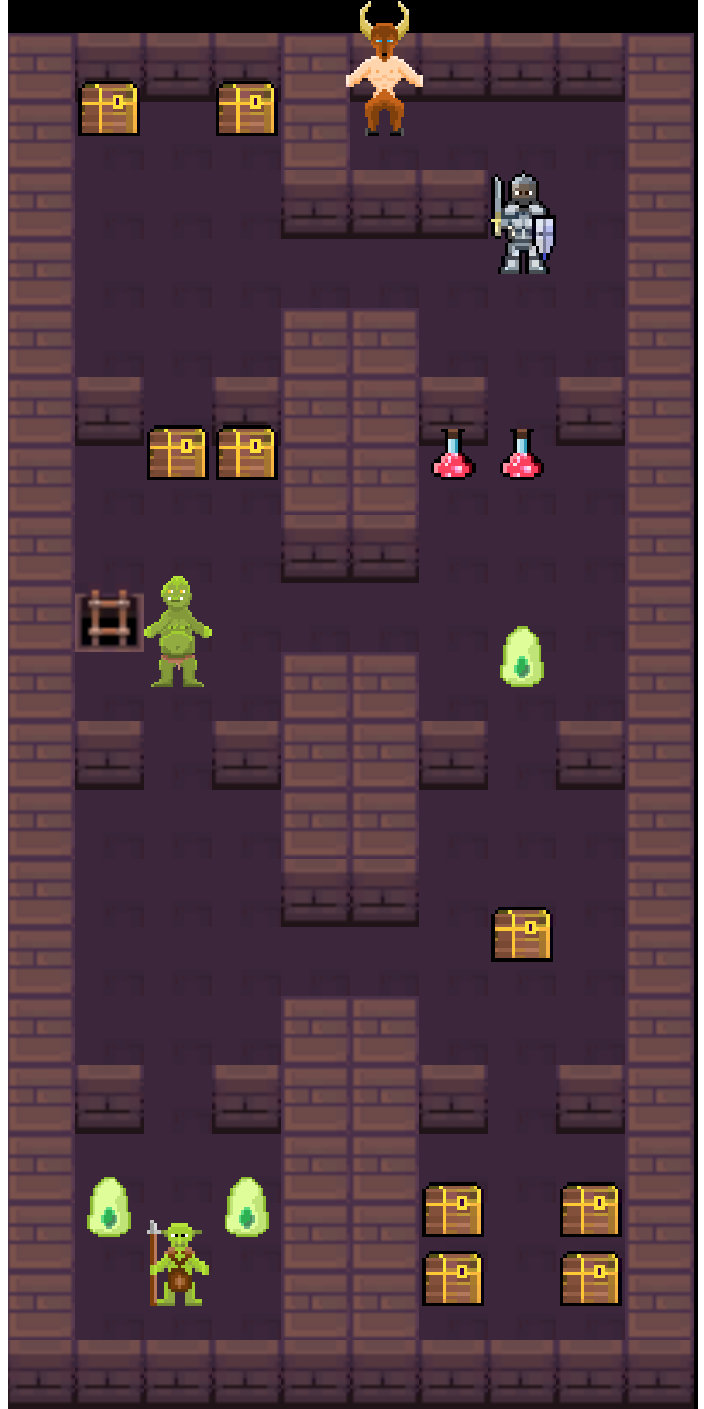}
    \includegraphics[height=40ex]{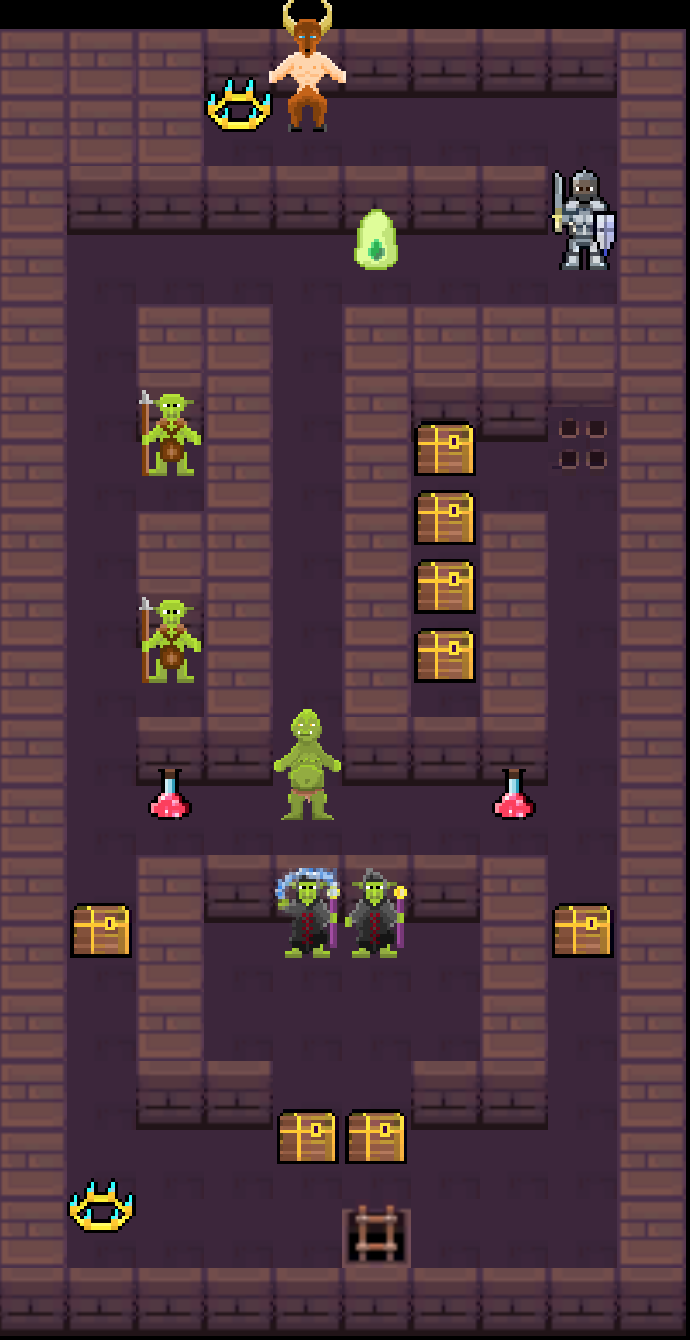}
    \caption{Human designed levels from MiniDungeons 2.}
    \label{fig:minidungeons}
    \vspace{-5mm}
\end{figure}

MiniDungeons 2~\footnote{http://minidungeons.com/} (MD2) is a 2-dimensional deterministic, turn-based, rogue-like game~\cite{holmgaard2015minidungeons}, in which the player explores dungeon levels, with the end goal of reaching the exit in each level. Levels are made up of a mix of impassible \textit{walls} and passable \textit{floors}. The player can interact with various game objects and enemies scattered throughout the level. To win, the player must reach the exit, represented as a staircase. All game characters have Hit Points (HP) and deal damage when the player collides with them. The player begins the game with 10 HP, and if the player runs out of HP, they die and lose. Figure~\ref{fig:minidungeons} shows a group of different MD2 levels that contain various different game objects.

Each turn, the player selects an action to perform, and all game characters will then move after the player completes their action. Any game character may move in one of the four cardinal directions (North, South, East, West) on their turn so long as the tile in that direction is not a wall. The player is given one re-usable javelin at the start of every level. The player may choose to throw this javelin and do 1 damage to any monster within their unbroken line of sight. After using the javelin, the hero must traverse to the tile to which it was thrown to pick it up in order to use it again.

Each level contains a mix of game objects . Each game object has a different effect upon interacting with it:
\begin{itemize}
\item\textbf{Potions} heals the hero's HP by 2, up to the maximum of 10.
\item\textbf{Treasures} increase the hero's score upon collecting.  
\item\textbf{Portals} come in pairs and they teleport any character from one location to the other on the same turn.
\item\textbf{Traps} deal 1 damage to any character moving through them.
\end{itemize}
Levels also contain enemies, all of which desire to attack the player:
\begin{itemize}
\item\textbf{Goblins} move 1 tile towards the player if within line of sight. They have 1 HP and deal 1 damage upon collision.
\item\textbf{Goblin Wizards} cast a 1 damage spell at the hero if they have line of sight within 5 tiles. If they are over 5 tiles but have line of sight, they will move 1 tile towards the player. Wizards have 1 HP and deal no damage on collision.
\item\textbf{Blobs} move 1 tile towards a potion or the hero within line of sight. A blob colliding with a potion or another blob enables it to level-up into a more powerful blob. Blob starts with 1 HP and does 1 damage and increase by 1 HP and 1 damage for every level up with max of 3 HP and 3 damage.
\item\textbf{Ogres} will move 1 tile towards either the hero or a treasure within line of sight. When an ogre collides with a treasure, they consume it. Ogres have 2 HP and deal 2 damage to anything they collide with.
\item\textbf{Minitaurs} always move 1 tile along the shortest path to the hero, regardless of line of sight. Collision with the minitaur deals 1 damage and stun it for 5 turns where the player can move through. A minitaur has no HP and is immortal.
\end{itemize}


\section{Persona-Driven Level Generator}\label{sec:cme}
We use the evolutionary search-based algorithm known as Constrained MAP-Elites (CME)~\cite{khalifa2018talakat} to generate levels. Similar to MAP-Elites, CME uses a multi-dimensional grid of cells to store the chromosomes instead of an actual population. Each chromosome is placed in the grid based on a set of distinct behavior characteristics, while its fitness determines if it will be saved or thrown away. The difference is that each cell in CME has three different types of chromosomes: elite chromosome, feasible chromosomes, and infeasible chromosomes (similar to FI2Pop algorithm\cite{kimbrough2008feasible}). Feasible and infeasible chromosomes are identified based on if they satisfy a group of constraints. If a chromosome satisfies these chromosomes, they are placed in the feasible population, otherwise they are placed in the infeasible one. The elite chromosome is the most fitted chromosome in the feasible population. Chromosomes can be moved between cells if their behavior characteristic shift and/or between different populations within their cell if they successfully outgrow their constraints or fail to do so.

The system's evolutionary pipeline consists of 5 distinct steps: initialization, evaluation, replacement, selection, and mutation. The system uses these steps in the following fashion to generate levels:
\begin{enumerate}
    \item Initialize a multi-dimensional grid of cells where each cell have 2 populations for the feasible and the infeasible chromosomes where each population of size ``MAP\_CELL\_SIZE''.
    \item Generate an initial population of chromosomes of size ``POPSIZE'' using the \textbf{Initialization} step.
    \item Evaluate the current population of chromosomes fitness and behavior characteristics using the \textbf{Evaluation} step. \label{steps:evaluate}
    \item Insert the current population of chromosomes into the multi-dimensional grid using the \textbf{Replacement} step.
    \item Generate a new population of chromosomes of size ``POPSIZE'' \label{steps:generate}
    \begin{enumerate}
        \item Select a chromosome from the grid using the \textbf{Selection} step.
        \item make a new copy of the selected chromosome using the \textbf{Mutation} step.
    \end{enumerate}
    \item Repeat steps \ref{steps:evaluate} to \ref{steps:generate} for number of iterations equal to ``ITERATIONS''.
\end{enumerate}
Chromosomes (game levels) are represented as two 2D array of size (``LEVEL\_WIDTH'' x ``LEVEL\_HEIGHT'') where each tile can be any of the possible tile types in MD2 (empty, solid, hero, potion, treasure, trap, portal, goblin, goblin wizard, blob, ogre, and minitaur). Table~\ref{tab:hyperparams} provides a description of all hyperparameters as they are referenced in the previous steps and the following sections as well as their experiment values.

\begin{table*}[ht]
\resizebox{\textwidth}{!}{%
\begin{tabular}{|l|l|l|l|}
\hline
Phase                       & Parameter         & Description & Value                                                                                       \\ \hline
\hline
\multirow{2}{*}{Setup} & BUCKET  & Bucket size of the MAP-Elites matrix    & 5              \\ \cline{2-4}
                       & PERSONAS   & Number of dimensions that is used for the MAP-Elites matrix  & 3  \\ \cline{2-4} 
                       & MAP\_CELL\_SIZE   & Maximum number of members allowed to be stored in a cell for the feasible/infeasible population  & 5  \\ \cline{2-4} 
                       & LEVEL\_WIDTH   & The width in tiles of evolved MD2 levels  & 10  \\ \cline{2-4} 
                       & LEVEL\_HEIGHT   & The height in tiles of evolved MD2 levels  & 10  \\ \hline
\multirow{2}{*}{Evolve}     & ITERATIONS        & How many times to evolve (evaluate, select, mutate) a population  & 500                    \\ \cline{2-4} 
                            & POPSIZE           & Maximum size of the population. Members may be removed if they are invalid to be evaluated      & 60   \\ \hline
\multirow{2}{*}{Initialize} & EMPTY\_INIT\_RATE & Probability to place an empty tile in the initialization phase of the map & 0.5                          \\ \cline{2-4} 
                            & WALL\_INIT\_RATE        & Probability to place a wall tile in the initialization phase of the map  & 0.3  \\ \cline{2-4} 
                            \hline
\multirow{2}{*}{Mutation}   & EMPTY\_MUT\_RATE  & Probability to place an empty tile in the mutation phase of the map  & 0.5                              \\ \cline{2-4} 
                            & MUTATION\_RATE    & The probability to mutate the current tile (iterates over every tile in the map excluding borders) & 0.1 \\ \hline
\multirow{2}{*}{Selection}  & ELITE\_PROB       & Probability to select a sample from the elite population     & 0.8                                      \\ \cline{2-4} 
                            & FEAS\_PROB        & Probability to select a sample from the feasible population   & 0.6                                   \\ \hline
\multirow{2}{*}{Personas}  & C       & Weight constant for agent's utility cost & 45                                      \\ \cline{2-4} 
                            & K        & Weight constant for agent's death cost  & 1                                   \\ \hline                            
\end{tabular}
}
\caption{Hyperparameter descriptions and experimental values}
\label{tab:hyperparams}
\vspace*{-5mm}
\end{table*}


\subsection{Initialization}\label{sec:initialization}
The system initially creates blank maps (empty map with only wall border) with a fixed width and height (``LEVEL\_WIDTH'' x ``LEVEL\_HEIGHT''). In our experiments, level width and height are fixed to a size of 10x10, in contrast to the traditional MD2 levels which are 10x20. We selected this smaller size as it allows for faster agent evaluation which is described in Section~\ref{sec:proc_persona}.

Every tile in the map (except the border tiles) is iteratively changed to a random tile based on predefined probabilities where empty and wall tiles have higher probabilities than other game elements (empty probability and wall probabilities are defined in Table~\ref{tab:hyperparams} as ``EMPTY\_INIT\_RATE'' and ``WALL\_INIT\_RATE'' respectively). After this step, the randomly initialized levels are repaired such that these levels contain exactly one player, one exit, and either exactly 2 or 0 portals. The algorithm repairs the generated levels by either placing the missing tile at a random location or removes the excessive tiles until the levels are fixed.

\subsection{Evaluation}\label{sec:evaluation}
The maps are placed into a Constrained MAP-Elites grid with each cell in the matrix containing two populations: feasible and infeasible. The feasibility of the map is determined by calculating a constraint value. The constraint value for the maps are based on the connectivity of the map. A breadth-first algorithm is conducted on every non-wall tile in the map starting from the player's initial position, and if every non-wall tile is reached the constraint value is 1 and the map is considered feasible. Infeasible maps contain disconnected non-wall tiles and the constraint value is based on the ratio between reachable tiles ($t_{reach}$) and all non-wall tiles ($t_{total}$) as shown in the following equation:
\begin{equation}\label{eq:constarint}
    S = \frac{t_{reach}}{t_{total}}
\end{equation}

The fitness function measure the simplicity of the levels. To calculate simplicity of the generated levels, we calculate the entropy of the different tiles in the generate level similar to Charity et al.~\cite{charity2020mech}. With this fitness, the generator will try to evolve levels that have small amount different tiles which makes levels look less noisy. Equation \ref{eq:entropy_fitness} displays the map fitness equation, where $p_{i}$ is the percentage of a specific tile $i$ existing on the level and $n$ is the total number of distinct tiles in MD2.
\begin{equation}\label{eq:entropy_fitness}
\begin{aligned}
    &fitness = -\sum_{i=0}^{n} p_{i} \cdot \log(p_{i})
\end{aligned}
\end{equation}

Finally, in order to place a newly generated level in the CME grid, their behavior characteristic need to be calculated. In this work, the behavior characteristics is the hero's remaining health after level completion by each persona, ranging from 0 (death) to 10 (full health). The health is divided into buckets of $10 / BUCKET$ health where ``BUCKET'' is the hyperparameter that determines the size of each dimension. With multiple different personas, the size of the MAP-Elites matrix will be $BUCKET^{PERSONAS}$ cells, where ``PERSONAS'' is a hyperparameter that determines the number of used personas. In our experiments, we use 3 different personas which are runner, monster killer, and treasure collector; and 5 buckets per dimension. More information about the different personas will be described in Section \ref{sec:proc_persona}.

\subsection{Replacement}
The elites of this matrix are defined as the levels with the highest fitness values in the feasible population of the cells. Each cell has a limit to the size of the population (``MAP\_CELL\_SIZE''). When a new generated level has a better fitness or constrained value than levels in the current cell's feasible or infeasible population, the least fit level in that population is removed from the group and the new level replaces it. In the case of the feasible population, if the new level has better fitness than the least fit level in this population, the new level is inserted and the least fitted level will be moved to the infeasible population instead of being completely deleted. On the other hand, if the new level is going to be inserted in the infeasible population, it has to be more fit than the least performing level to replace it. After a couple of iterations, the infeasible population will contain substantially fewer maps with low constraint values as more fitted individuals are being pushed to it. This "survival of the fittest" replacement for the feasible and infeasible populations maintains the quality of the matrix and all of the maps contained in the cells while providing better selection possibilities for future populations.

\subsection{Selection}
In the selection step, it is important to note that behavioral characteristics are not considered at all. Every elite level from the matrix is added into an elite selection pool indiscriminately. Members of the feasible and infeasible populations are added to their own respective pools. Elites are given the highest probability (``ELITE\_PROB'') to be selected, then feasible (``FEAS\_PROB''), and lastly infeasible ($1-ELITE\_PROB-FEAS\_PROB$). When a set is chosen, a chromosome is selected randomly from the set to be mutated (explained in Section~\ref{sec:mutation}) before being added to the next generation's population.

\subsection{Mutation}\label{sec:mutation}
During the mutation step, every tile of a selected map is given a chance of mutating as controlled by the rate of tile mutation (``MUTATION\_RATE''). If a tile is set to mutate, it may change itself into any tile in the game with equal probability except for an empty tile, which has a higher chance to be selected as the end mutation (``EMPTY\_MUT\_RATE''). All the other tiles have equal probability to be selected. We have the empty tile with higher probability than the rest to encourage the evolution to erase more often than adding. The same repair function from the initialization phase is used to ensure there is only 1 player, 1 exit, and 2 or 0 portals (see Section \ref{sec:initialization}). Additionally, If a mutated level caused the player to not be able to physically reach the exit. The level will be removed and won't be inserted back to the CME grid as these levels don't allow agents to determine their behavior characteristics (as they are not winnable). Consequentially, an iteration may add less chromosomes than the predefined size (``POPSIZE''), however all of the levels will be guaranteed to be playable.

\section{Procedural Personas}\label{sec:proc_persona}
Our experiments use an online-planning A* agent to evaluate levels. Each turn, the agent is limited to building a $500$ node tree to decide their next action. $500$ nodes is chosen after preliminary experiments suggest that an A* agent can play well but still be deceived with the resulting limited planning horizon. In theory, this could be varied to approximate player skill~\cite{isaksen2017simulating}, however we do not perform that in this paper. The A* agent plays every level three times using three different playstyles (personas). The results of these three runs determine that level's behavioral characteristics. A* does not need to play the level more than once per persona due to the deterministic nature of both the game and the A* algorithm.

As mentioned in section~\ref{sec:evaluation}, the behavior characteristics of a level correspond to the remaining health of the different personas after completing the tested level. In previous work by Holmg{aa}rd et al.~\cite{holmgaard2014personas,holmgaard2018automated}, three main personas are identified for MD2:
\begin{enumerate}
    \item \textbf{Runner (R):} complete the level as fast as possible.
    \item \textbf{Monster Killer (MK):} slay as many monsters as possible before completing the level.
    \item \textbf{Treasure Collector (TC):} open as many chests as possible before completing the level.
\end{enumerate}
Each of these personas use a utility function which influences their behavior. Depending on the persona, the agent uses one of the following heuristics and cost functions. In the following equations, $h_{persona}$ and $g_{persona}$ denote the heuristic function and cost function respectively for a specific player persona ($persona$). The personas are delineated as runner ($r$), monster killer ($mk$), or treasure collector ($tc$).

A Runner agent (R) tries to get to the exit in the fewest amount of steps as possible. Equation~\ref{eq:r} displays the heuristic ($h_{r}$) and cost ($g_{r}$) function of the runner agent.
\begin{equation}\label{eq:r}
    \begin{aligned}
   &h_{r} = dist_{exit}\\
   &g_{r} = -steps
    \end{aligned}
\end{equation}
\noindent where  $dist_{exit}$ is the distance from the current player location to the exit, and $steps$ is the amount of steps taken since the start.

A Monster Killer agent (MK) tries to find the shortest path to the closest monster, killing all monsters while attempting not to die, and getting to the exit when all monsters have been slain. Equation~\ref{eq:mk} shows the heuristic ($h_{mk}$) and the cost ($g_{mk}$) functions of the monster killer agent.
\begin{equation}\label{eq:mk}
    \begin{aligned}
   &h_{mk} = \begin{cases} 
     min(dist_{monster}) & N_{monster} > 0 \\
     dist_{exit} & N_{monster} == 0
      \end{cases}\\
   &g_{mk} =  c \cdot N_{monster} + k \cdot Dead
    \end{aligned}
\end{equation}
\noindent where $c$ and $k$ are constants (see Table~\ref{tab:hyperparams} for used values), $dist_{exit}$ is the distance between the player and the exit, $min(dist_{monster})$ is the distance to the closest monster from the player location,  $N_{monster}$ is the number of alive monsters in the level, and $Dead$ is a binary value that is equal to 1 if the player is dead and 0 otherwise.

A Treasure Collector agent (TC) tries to find the shortest path to the nearest treasure, collecting all treasures while attempting not to die, and getting to the exit when all treasures have been collected. It uses a similar equation to~\ref{eq:mk} but replacing $min(dist_{monster})$ with $min(dist_{treasure})$ (the distance to the closest treasure from the player location) and $N_{monster}$ with $N_{treasure}$ (number of unopened treasures in the level).

\section{Results}\label{sec:results}
In this section, we review the results from CME over five runs. When we refer to a level's behavioral characteristics, we refer to its bucketed values in order of runner, treasure collector, and monster killer HP results. Buckets are numbered 0 to 4, totaling 5 buckets. For example, a level with dimensions of 444 means that the agent finished the level in the 4th bucket (the highest amount of HP) with every persona. We divide the levels into 3 major types:
\begin{itemize}
    \item \textbf{Balanced:} These are levels where all the three personas complete the level with approximately the same HP. These are the diagonal cells in the CME matrix (000, 111, 222, 333, and 444).
    \item \textbf{Dominant:} These are levels where a certain persona (R, TC, or MK) completes the level with more HP than the other two personas. We call them Runner dominant, Monster Killer dominant, and Treasure Collector dominant levels.
    \item \textbf{Submissive:} These are levels where a certain persona (R, TC, or MK) completes the level with less HP than the other two persona. We call them Runner submissive, Monster Killer submissive, and Treasure Collector submissive levels. 
\end{itemize}
By differentiating levels this way, we can better analyze a level's ability to encourage or discourage a player to behave as a specific persona. First, we analyze the resulting MAP-Elite matrix, focusing on the elite coverage. We then analyze the levels in terms of their persona HP outcomes, as defined in the paragraph above.

\subsection{Elite Matrix}
\begin{figure}
    \centering
    \includegraphics[width=\columnwidth]{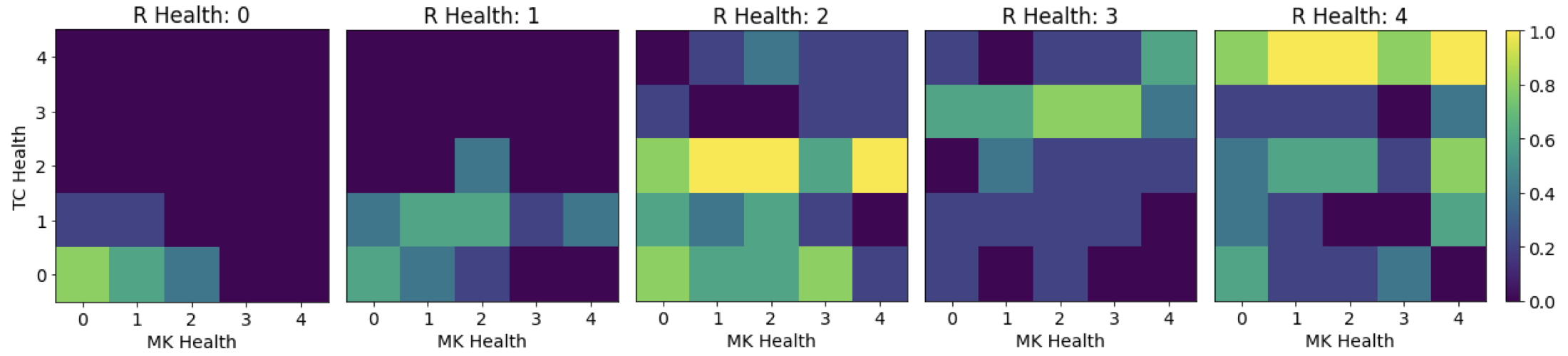}
    \caption{The probability of an elite being found for each cell over 5 different runs.}
    \label{fig:elites_map}
\end{figure}

\begin{table}
    \centering
    \resizebox{\columnwidth}{!}{%
    \begin{tabular}{|c||c|c||c|}
        \hline
        Persona & Dominant & Submissive & \cellcolor[gray]{0.8}Max Levels \\
        \hline
        \hline
        Runner & \textbf{9.0 ± 3.16} &  2.0 ± 1.67 & \cellcolor[gray]{0.8}30 \\
        \hline
        Treasure Collector & 1.8 ± 0.75  & \textbf{7.4 ± 1.74} & \cellcolor[gray]{0.8}30  \\
        \hline
        Monster Killer & 6.0 ± 2.45 & \textbf{11.8 ± 2.31} & \cellcolor[gray]{0.8}30 \\
        \hline
        \hline
        Balanced & \multicolumn{2}{c||}{\textbf{4.2 ± 0.75}} & \cellcolor[gray]{0.8}5 \\
        \hline
    \end{tabular}
    }
    \caption{Mean and standard deviation of the number of different level types discovered across 5 runs.}
    \label{tab:level_stats}
    \vspace{-7mm}
\end{table}

The elite map contains 5x5x5 cells, or $125$, since each dimension is made of $5$ buckets. If we aggregate the five runs, CME fills an average $34.2$ cells ($27.36\%$) with standard deviation of $4.82$. Figure~\ref{fig:elites_map} displays the elite map of populated cells across 5 experimental runs. Higher runner health seems to correlate to higher cell coverage. This is likely to be because monster killer and treasure collector heuristics will default to runner heuristics (get to the exit) once their primary objectives are completed. Generating a level where a runner persona loses health while other algorithms loses less health in comparison is difficult to do, as it is not easy to deceive a runner without also deceiving the other two personas. The runner's final health tends to act as an upper bound for the final health of the other agents. Although there are maps where TC and MK personas complete the level with higher health, they are less likely to be found.

To understand more about the ability of CME to generate different types of levels, we analyze the number of \emph{dominant} and \emph{submissive} level types across 5 experimental runs as shown in table~\ref{tab:level_stats}. The runner persona tends to dominate the other two personas more often than it is submissive. As mentioned before, both treasure collector and monster killer personas behave similar to the runner when there are no treasures or monsters respectively. In contrast, balanced levels seem relatively easy to discover, as on average 4 levels are found out of a maximum of five per experimental run. If both treasure and monsters are placed along or nearby the shortest path to the exit, all three personas tend to behave the same and therefore have identical HP outcomes. It is also relatively simple to create impossible levels by placing lots and lots of monsters near the exit (see Figure \ref{fig:balanced_hard_levels}).

\subsection{Balanced Levels}
\begin{figure}
    \centering
    \begin{subfigure}[t]{\columnwidth}
        \centering
        \includegraphics[width=\textwidth]{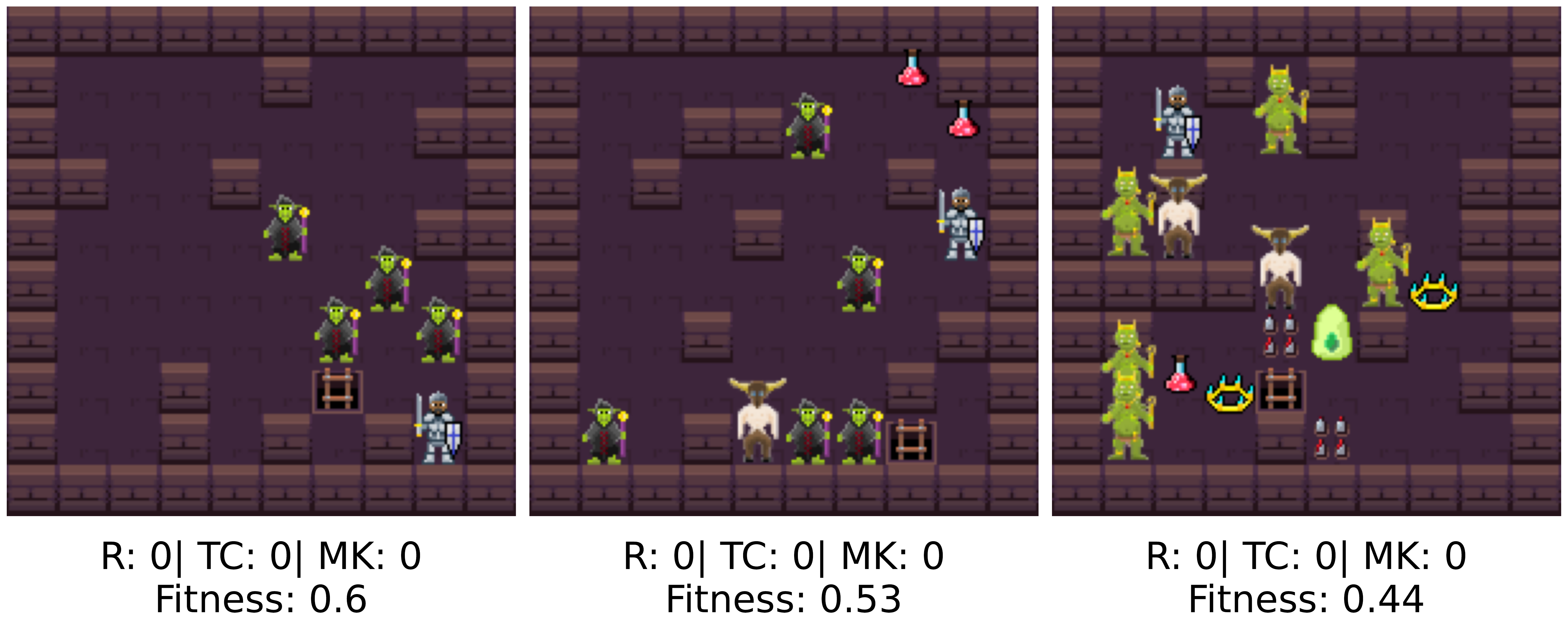}
        \caption{All personas die/nearly die}
        \label{fig:balanced_hard_levels}
    \end{subfigure}
    \begin{subfigure}[t]{\columnwidth}
        \centering
        \includegraphics[width=\textwidth]{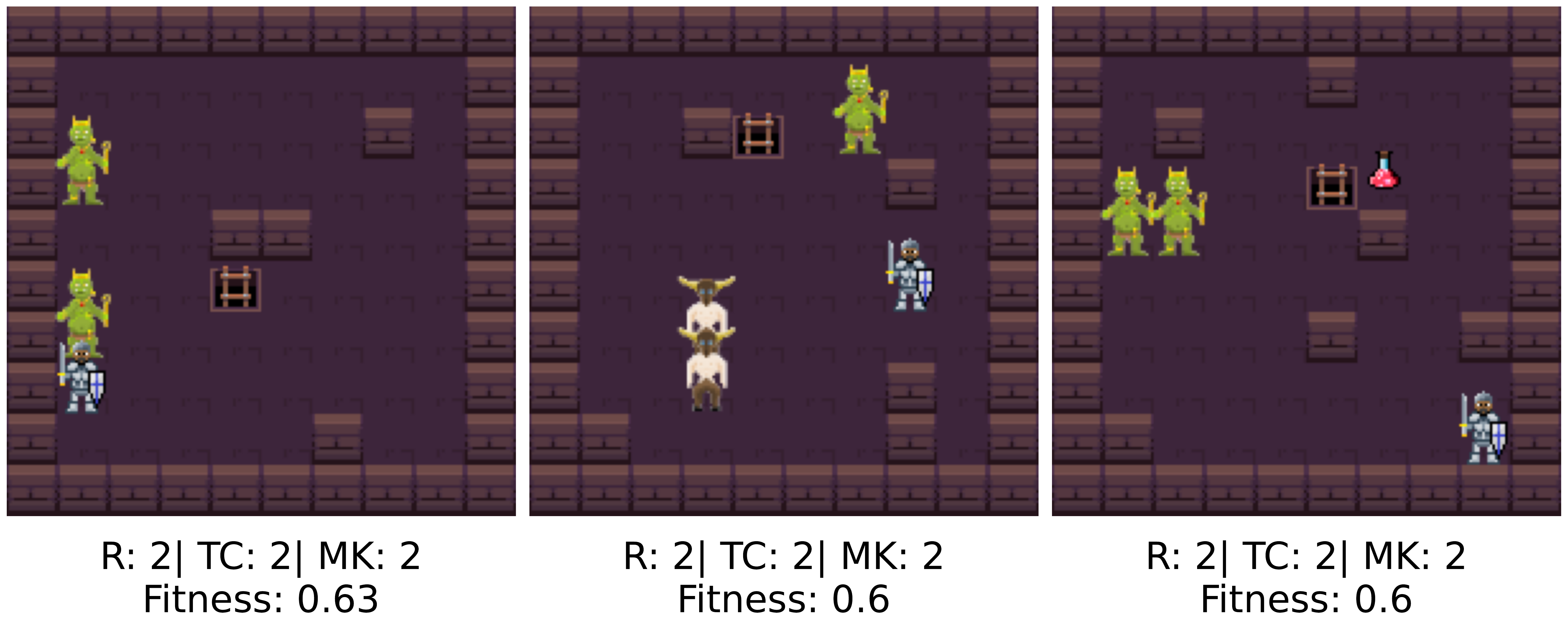}
        \caption{All persona lose some health}
        \label{fig:balanced_medium_levels}
    \end{subfigure}
    \begin{subfigure}[t]{\columnwidth}
        \centering
        \includegraphics[width=\textwidth]{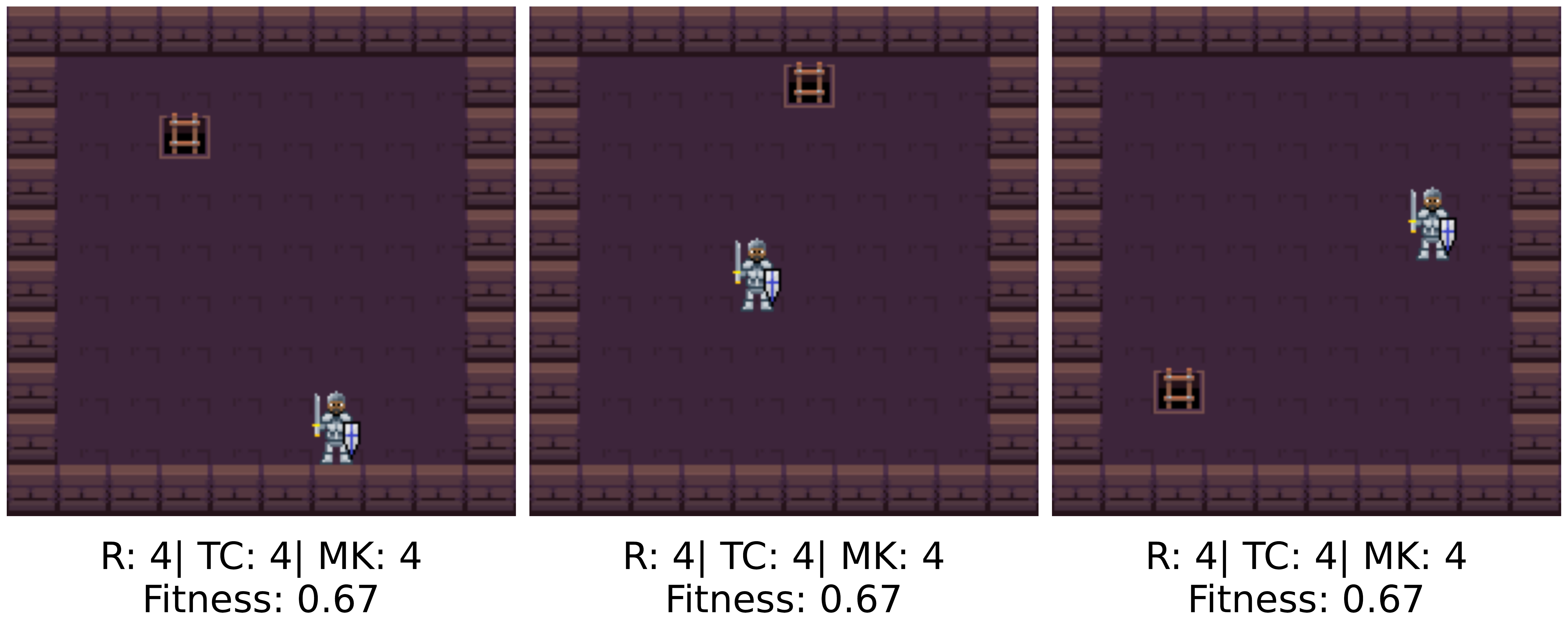}
        \caption{No personas lose health}
        \label{fig:balanced_easy_levels}
    \end{subfigure}
    \vspace{-2mm}
    \caption{Simplest generated balanced levels across 5 runs.}
    \label{fig:balanced_levels}
    \vspace{-5mm}
\end{figure}
Balanced levels are the levels in which all three personas finish with a similar health percentage. These levels have behavior characteristics of either 000, 111, 222, 333, or 444. A level with a behavior characteristic of 444, means that the agent finished the level with at least 8 or more HP remaining with every persona.

Figure \ref{fig:balanced_levels} displays 3 example levels for 3 different balanced levels values: 000, 222, and 444. The 444 levels (Figure~\ref{fig:balanced_easy_levels}) are levels in which all 3 personas finish with nearly all of their HP. These are the simplest, easiest possible levels, consisting of just the hero and the exit. The only difference between these levels is the position of the player and exit location. Since the system has no constraint on the solution length, a level where the player starts next to an exit is scored similar to a level that requires the player to learn how to move and navigate a maze. However, the simplicity-based fitness function will pressure the generator to evolve empty levels since they will have higher fitness. We did not include path length as part of the constraints or fitness, since we wanted these levels to encourage a user to play as a certain persona and not how to move and play in general. Figure~\ref{fig:balanced_hard_levels} displays levels where all three personas die or have very low health (000 levels). Placing lots of enemies along the shortest path towards the exit quickly obliterates HP regardless of the persona. Figure~\ref{fig:balanced_medium_levels} displays levels where all 3 persona have medium amount of health (between 4 to 6 HP). Levels such as these tend to have a small amount of monsters that will deal some damage to the player. Similar to the hard levels in Figure \ref{fig:balanced_hard_levels}, the enemies are close to shortest path towards the exit. 

\subsection{Dominant Levels}

\begin{figure}
    \centering
    \begin{subfigure}[t]{\columnwidth}
        \centering
        \includegraphics[width=\textwidth]{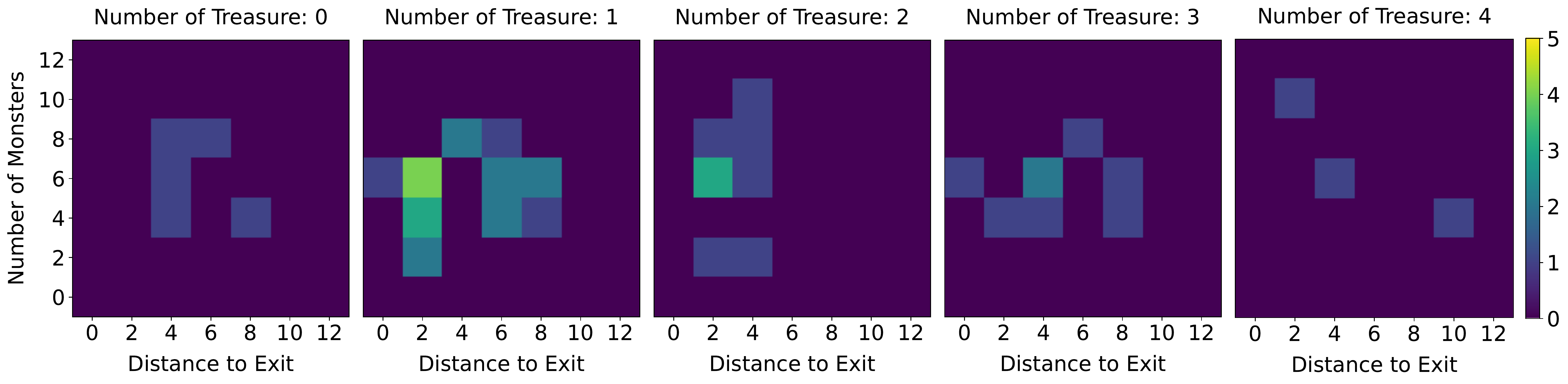}
        \caption{Runner Dominant Levels}
        \label{fig:r_dom_exp}
    \end{subfigure}
    \begin{subfigure}[t]{\columnwidth}
        \centering
        \includegraphics[width=\textwidth]{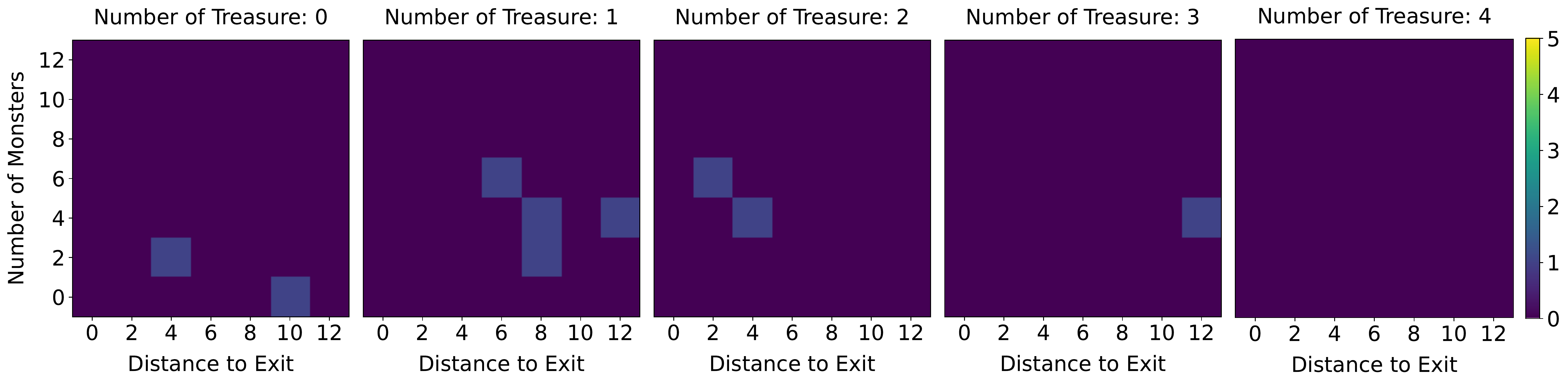}
        \caption{Treasure Collector Dominant Levels}
        \label{fig:tc_dom_exp}
    \end{subfigure}
    \begin{subfigure}[t]{\columnwidth}
        \centering
        \includegraphics[width=\textwidth]{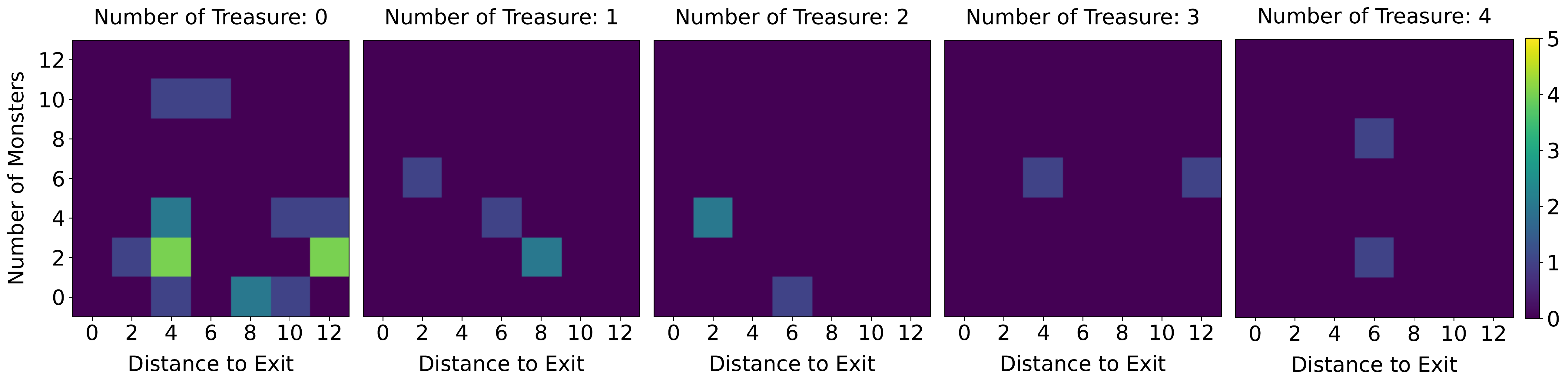}
        \caption{Monster Killer Dominant Levels}
        \label{fig:mk_dom_exp}
    \end{subfigure}
    \vspace{-2mm}
    \caption{Expressive range analysis for the different dominant levels across 5 runs.}
    \label{fig:dom_expressive}
    \vspace{-4mm}
\end{figure}

\begin{figure}
    \centering
    \begin{subfigure}[t]{\columnwidth}
        \centering
        \includegraphics[width=\textwidth]{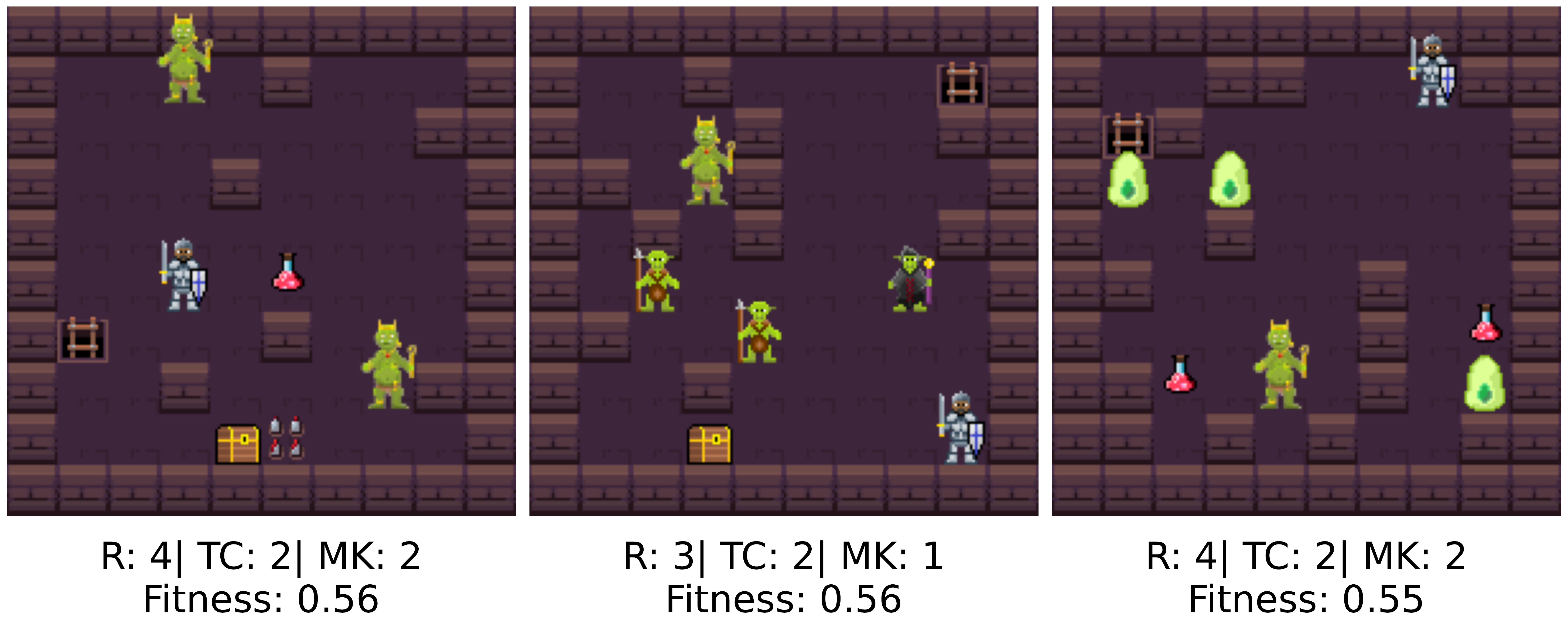}
        \caption{Runner Dominant Levels}
        \label{fig:r_dom_level}
    \end{subfigure}
    \begin{subfigure}[t]{\columnwidth}
        \centering
        \includegraphics[width=\textwidth]{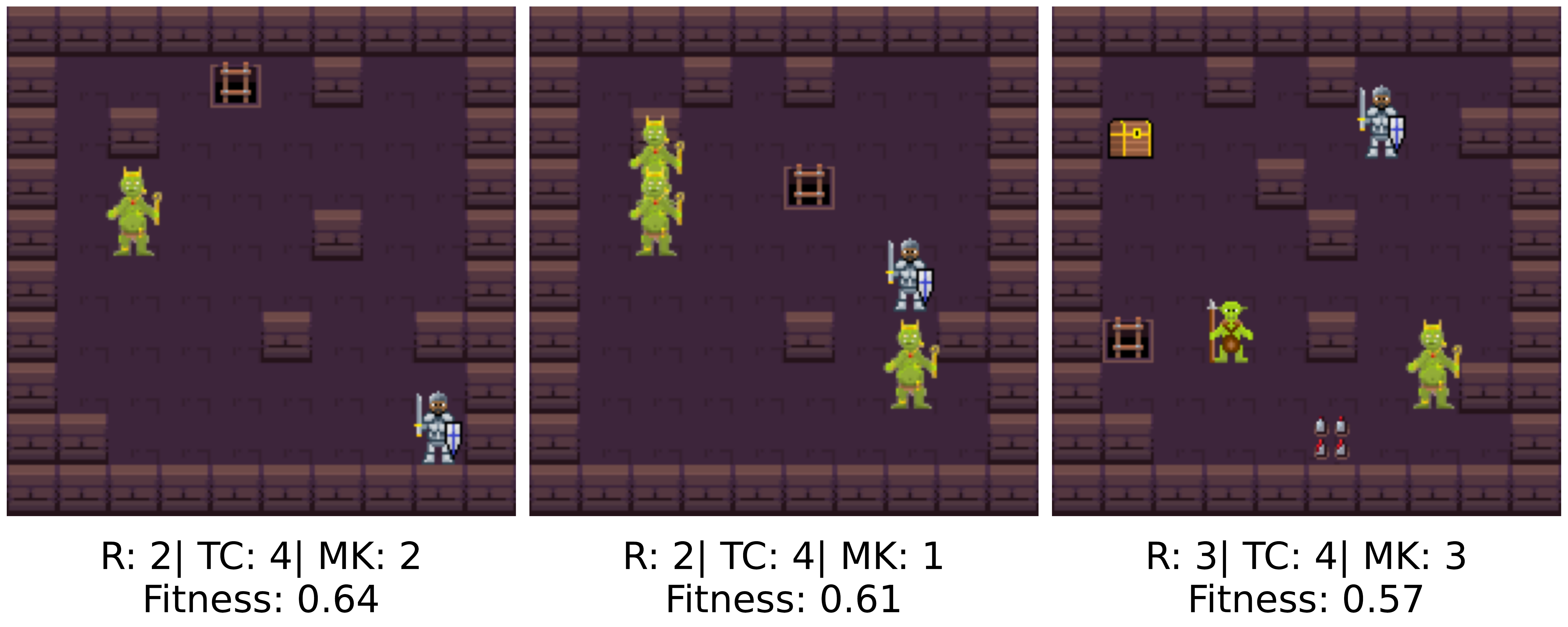}
        \caption{Treasure Collector Dominant Levels}
        \label{fig:tc_dom_level}
    \end{subfigure}
    \begin{subfigure}[t]{\columnwidth}
        \centering
        \includegraphics[width=\textwidth]{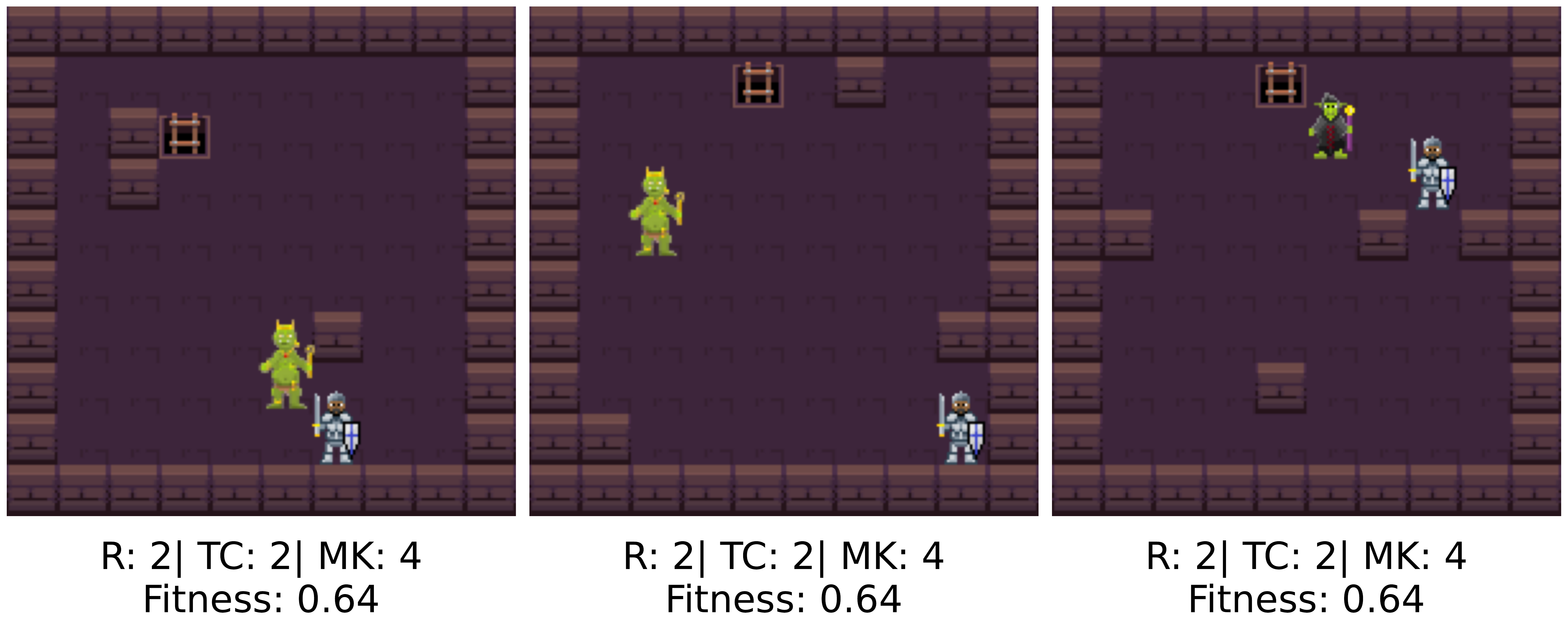}
        \caption{Monster Killer Dominant Levels}
        \label{fig:mk_dom_level}
    \end{subfigure}
    \vspace{-2mm}
    \caption{Simplest generated dominant levels where a certain persona ends with higher level than the other two.}
    \label{fig:dom_levels}
    \vspace{-5mm}
\end{figure}

Dominant levels are levels where one persona completes the level with higher HP than the other two personas. For example, a 321 level is a runner dominant level, while a 242 level is a treasure collector dominant level, and a 024 level is a monster killer dominant level. Since the dominant levels cover 25\% (30 cells out 125 cells as shown in table~\ref{tab:level_stats}) of the map, we will focus on the dominant levels that have the highest fitness in the elite map. Figure~\ref{fig:dom_expressive} displays expressive range analysis of all the discovered dominant levels across 5 runs. Runner dominant levels (Figure~\ref{fig:r_dom_exp}) tend to have at least 1 treasure or more and consistently have short distances to the exit. Monster killer dominant levels (Figure~\ref{fig:mk_dom_exp}) have longer paths to exit with fewer monsters than the runner dominant levels. There are not too many treasure collector dominant levels as they appear to be difficult to discover (only 9 levels across 5 runs). We believe this is because of the difficulty of using treasures to guide the treasure collector down a different path where they do not take damage while the runner does.

Figure~\ref{fig:dom_levels} show the highest fittest levels where runner, treasure collector, and monster killer dominates respectively. The runner dominant levels (Figure~\ref{fig:r_dom_level}) contain short direct paths to the exit with a few enemies nearby. A treasure chest is sometimes placed close to the rest of the monsters so that the treasure collector takes damage from the surrounding monsters on its way back to exit after collecting the treasure. Occasionally there is no treasure but still the treasure collector lose health, due to the different cost functions between the personas influencing the agent to take different paths. The monster killer dominant levels always have one enemy beside the shortest path to the exit so the monster killer can just kill it using the javelin without taking any damage while the other two personas encounter them. The treasure collector dominant levels (Figure~\ref{fig:tc_dom_level}) usually have treasures around the map to guide the treasure collector persona away from monsters (see Figure \ref{fig:tc_dom_exp}). We show two of the simplest maps that do not have treasure yet the treasure collector manages to avoid taking much damage anyway, again due to the nuances of the collector's heuristic and cost functions. We did not expect this type of behavior from treasure collectors, and we want to review the heuristic and cost functions of collectors in future work in order to generate more treasure collector dominant levels with treasures in them.


\subsection{Submissive Levels}

\begin{figure}
    \centering
    \begin{subfigure}[t]{\columnwidth}
        \centering
        \includegraphics[width=\textwidth]{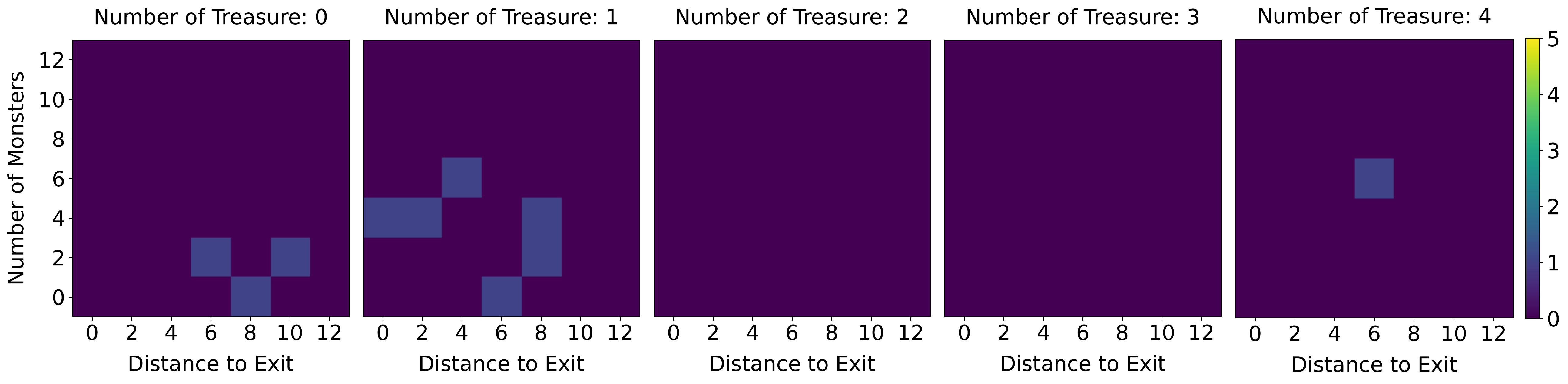}
        \caption{Runner Submissive Levels}
        \label{fig:r_sub_exp}
    \end{subfigure}
    \begin{subfigure}[t]{\columnwidth}
        \centering
        \includegraphics[width=\textwidth]{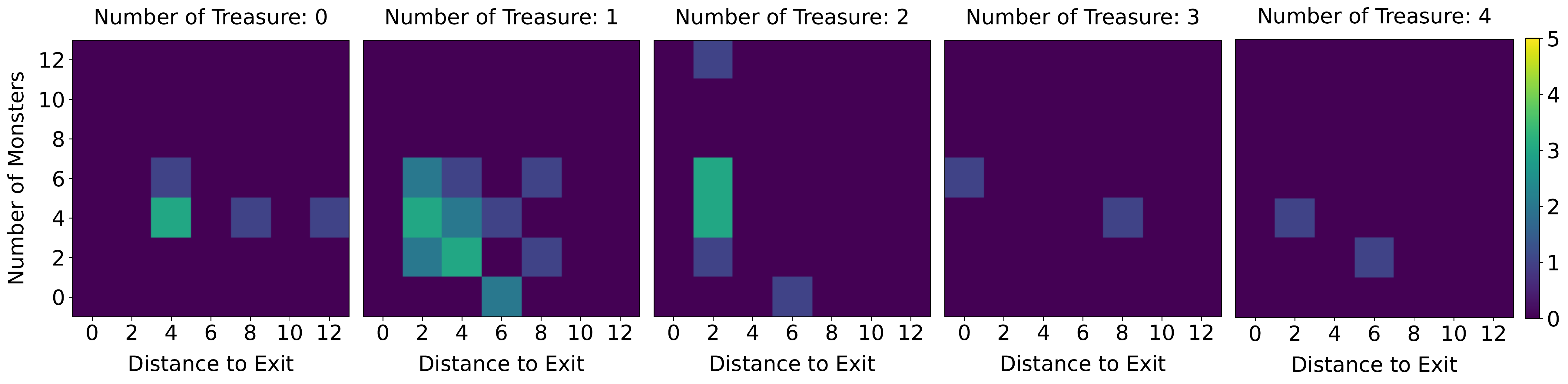}
        \caption{Treasure Collector Submissive Levels}
        \label{fig:tc_sub_exp}
    \end{subfigure}
    \begin{subfigure}[t]{\columnwidth}
        \centering
        \includegraphics[width=\textwidth]{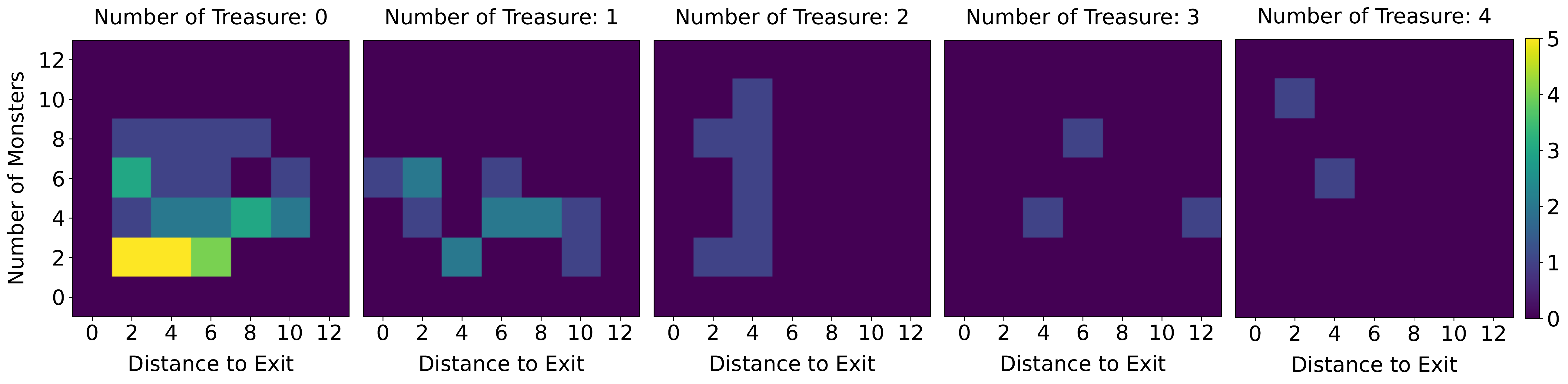}
        \caption{Monster Killer Submissive Levels}
        \label{fig:mk_sub_exp}
    \end{subfigure}
    \caption{Expressive range analysis for the different submissive levels across 5 runs.}
    \label{fig:sub_expressive}
    \vspace{-1mm}
\end{figure}

\begin{figure}
    \centering
    \begin{subfigure}[t]{\columnwidth}
        \centering
        \includegraphics[width=\textwidth]{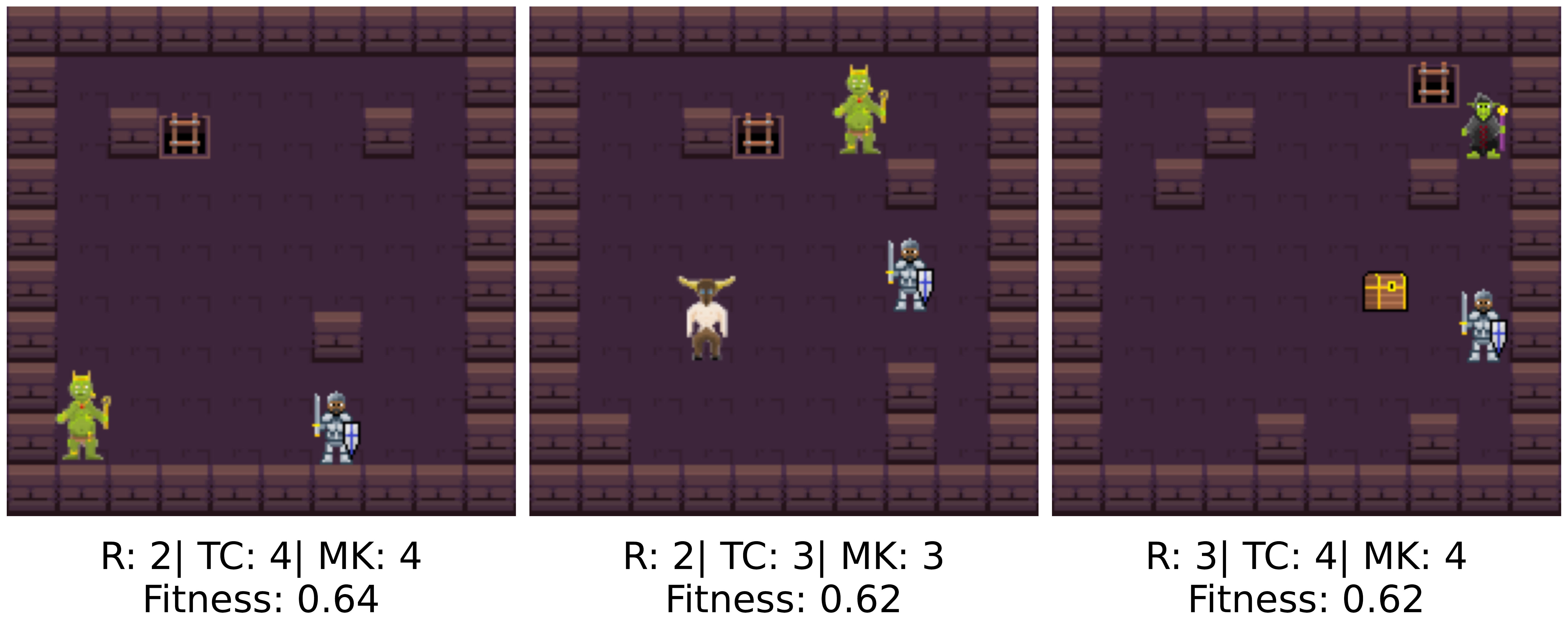}
        \caption{Runner Submissive Levels}
        \label{fig:r_sub_level}
    \end{subfigure}
    \begin{subfigure}[t]{\columnwidth}
        \centering
        \includegraphics[width=\textwidth]{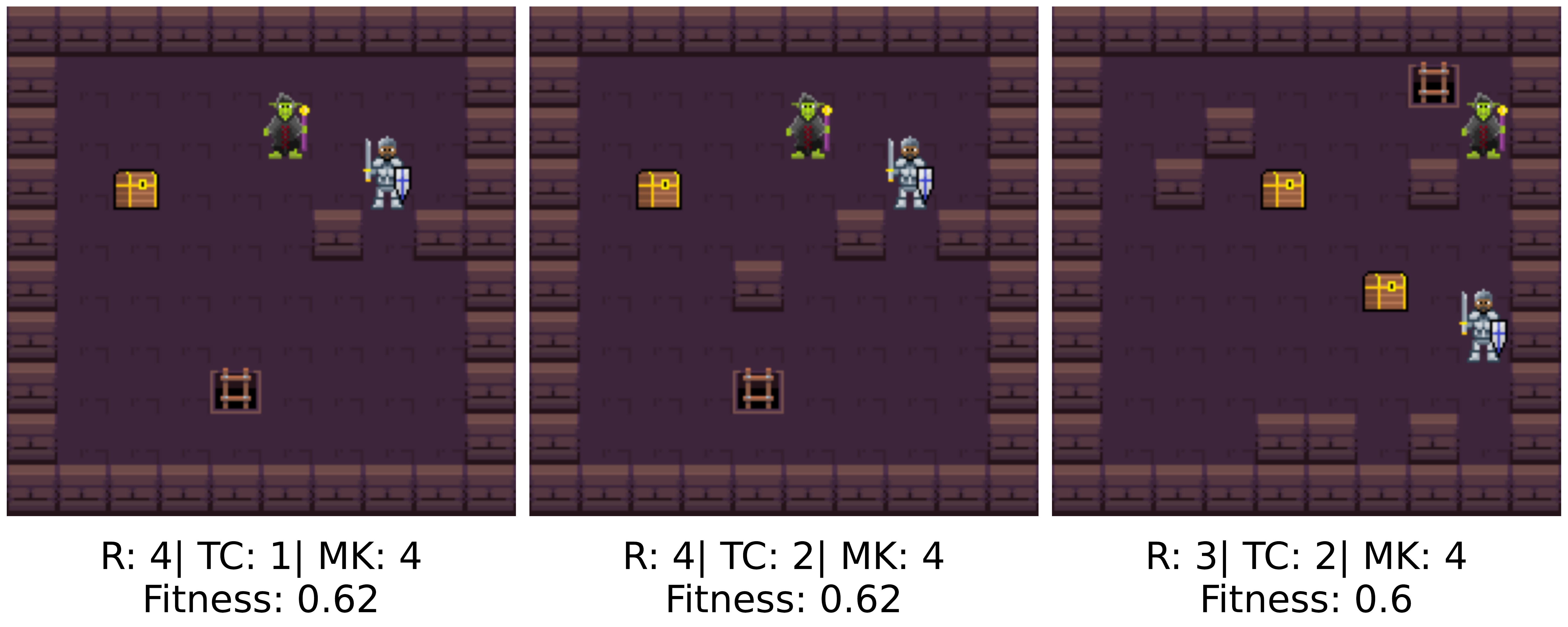}
        \caption{Treasure Collector Submissive Levels}
        \label{fig:tc_sub_level}
    \end{subfigure}
    \begin{subfigure}[t]{\columnwidth}
        \centering
        \includegraphics[width=\textwidth]{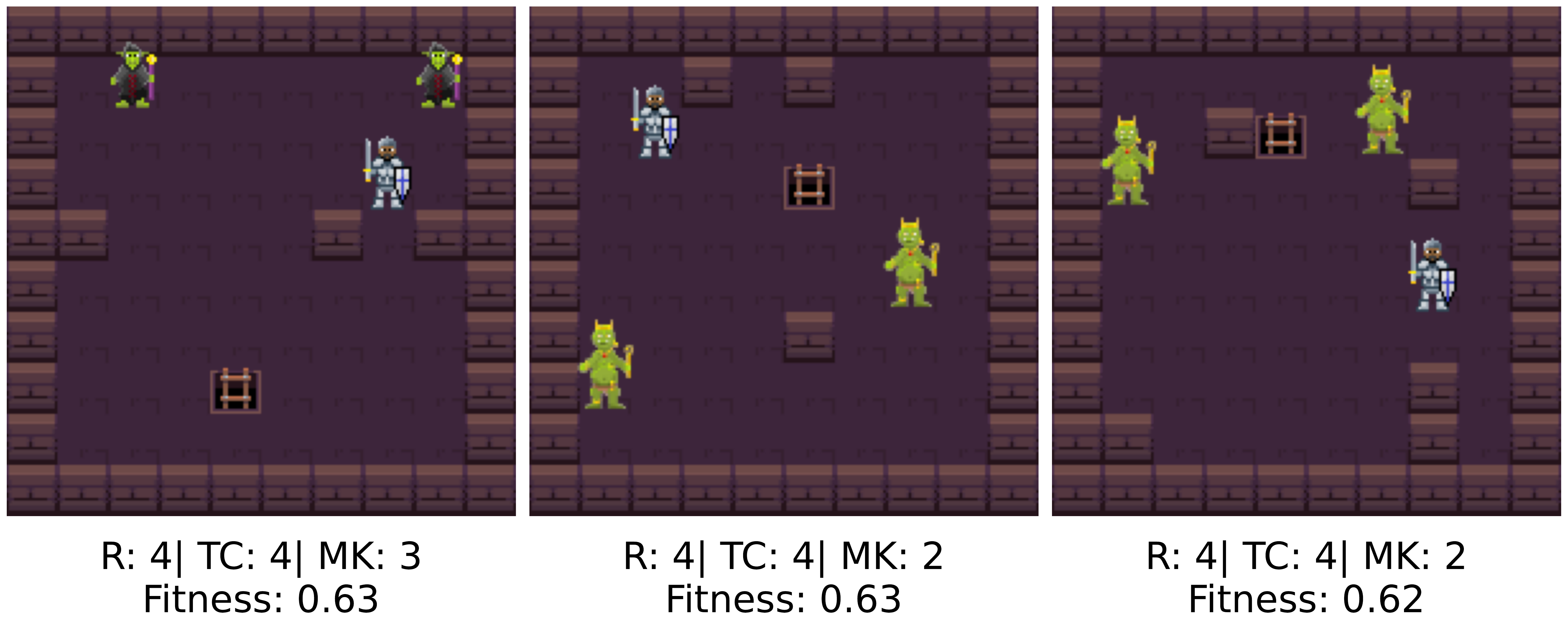}
        \caption{Monster Killer Submissive Levels}
        \label{fig:mk_sub_level}
    \end{subfigure}
    \vspace{-1mm}
    \caption{Simplest generated submissive levels where a specific persona finishes with lower health than the other two.}
    \label{fig:sub_levels}
    \vspace{-5mm}
\end{figure}

Submissive levels are levels where one of the personas finishes with lower HP than the other two. For example, a 143 level is a runner submissive level, a 301 level is a treasure collector submissive level, and a 241 is a monster killer submissive level. Figure~\ref{fig:sub_expressive} displays the expressive range analysis for the submissive levels. Runner submissive levels (Figure~\ref{fig:r_sub_exp}) are difficult to find (10 levels across 5 runs), which is expected as it is difficult to make a runner persona lose health without the other personas also losing health (monster killers especially). Treasure collector submissive levels (Figure~\ref{fig:tc_sub_exp}) tend to have at least one treasure chest, usually in proximity of monsters who attacks the nearby collector. Also, these levels usually have a short path to the exit allowing runner to run to exit with less taken damage. Monster killer submissive levels (Figure~\ref{fig:mk_sub_exp}) tend to have 0 or 1 treasures and small amount of monsters with short to decent distance to exit. Not having a lot of treasures influences the collector and runner personas to move towards the exit without battling with monsters while monster killer will always go to battle them and lose health. 

Similar to the dominant levels, we show the highest fitness level for each submissive persona in Figure~\ref{fig:sub_levels}. Looking at table~\ref{tab:level_stats} and figure~\ref{fig:sub_expressive}, monster killer or treasure collector submissive levels seem to be easier to discover than runner submissive ones. The treasure collector submissive levels (figure~\ref{fig:tc_sub_level}) place treasure and wizards in such a way that the runner can go directly to the exit without taking damage while the treasure collector gets hit by a wizard. In the runner submissive level (figure~\ref{fig:r_sub_level}), an enemy is placed close to the shortest path to the exit so it damages the runner on its way to the exit. Sometimes treasure is placed so that a treasure collector is guided away from the shortest path. The monster killer submissive levels (Figure~\ref{fig:mk_sub_level}) place monsters in such a formation that the runner and treasure collector do not have to engage them to win the level. The monster killer will engage them and take damage.

\section{Discussion}
In this section, we discuss the results through the lens of the motivation for this paper: tutorial levels. Our original goal is to generate levels which encourage or discourage a player in following the behavioral patterns of a certain persona. Below, we analyze levels that spotlight situations in which some persona tactics end in success and others more likely failure. We also discuss the correlation of behavioral characteristics and difficulty, arguing a level's difficulty is not determined solely from behavioral characteristics but that it does play a role.

\subsection{Tutorial Levels}
A notable subset of elites --- especially levels where the runner finishes with the most health out of the 3 personas --- generated levels that placed the player right next to the exit but had multiple monsters and/or treasures placed throughout the map (see Figure \ref{fig:r_dom_level}). These levels present a great example of encouraging certain playstyles for players - particularly for the runner persona. The monster killers will attempt to kill all of the monsters on the map and the treasure collector will try to get all of the treasure. But because of the overwhelming number of enemies, both personas either make it to the exit with very low health or die. However, the runner persona will only have to move a few steps to reach the exit, typically without facing any danger from traps or monsters. While these levels may not necessarily kill a player who plays however they want, they show that not every persona's strategy is easy, and they encourage a specific playstyle while punishing others.


Some of the levels encourage the player to use specific mechanics. For example, the monster killer dominant level shown in figure~\ref{fig:mk_dom_level} nudges the player towards throwing the javelin to avoid losing health. These levels may encourage the player to perform these mechanics but does not outright force them. If a designer wanted to force mechanic usage, they could simply change the player starting health. For example, if the player begins with only 2 HP in the monster killer dominant level (Figure~\ref{fig:mk_dom_level}, then the level will force the player to learn that the only way to win is to throw the javelin.

Similarly, submissive levels are good at encouraging players to avoid certain behaviors. For example, treasure collector submissive levels (Figure~\ref{fig:tc_sub_level}) may teach the player that not every treasure is safe to acquire since that there might be enemies protecting it. We can again make any submissive level an extreme variant by decreasing the player's starting HP to make these situations even more dangerous. The player will either need to learn to kill the enemies with the javelin to acquire the treasure (learning to be a monster killer) or avoid the treasure and run to the exit (learning to be a runner).

\subsection{Correlation to Difficulty}
Having a large amount of HP leftover for a persona does not always necessarily mean the level is ``easy'' for that persona. It does, however encourage the player to adopt the mindset of that persona style in order to maximize their ending HP. On the other hand, having a small amount (or hardly any) HP leftover for a persona typically does mean that playing as that persona is difficult.

One way to think about HP in MD2 is as a currency the player can spend. Some levels may require a player to spend more HP than others. On levels that require more spending, the player has less leeway in choosing when and how to spend that currency. In rogue-like game design literature, this concept is often referred to as ``strategic headroom''~\cite{smith2006headroom}. Typically levels that have more headroom are easier for the player to win (more options for the player to use) than levels that have less headroom. In MD2, levels that require less HP spending can offer the player more options to take (Figure~\ref{fig:balanced_medium_levels}) and therefore provide more headroom. Levels like the ones in Figure \ref{fig:balanced_hard_levels} afford very little headroom and are therefore very difficult.

\section{Conclusion}
This paper describes the development and results of a persona-driven tutorial level generation system that uses quality diversity in order to find and improve levels catered towards a specific playstyle. A set of generated elite levels are found in the solution space which address a broad range of playstyles. These levels can encourage a player towards trying a specific strategy ranging in difficulty or allow multiple playstyle types to succeed in a single level through their own strategy.  While this system uses the domain of MD2, a deterministic dungeon crawler game, our pipeline could be applied to other games so long as it has a clearly defined mechanic space allowing for a range of different playstyles. Our intention is to incorporate this system into a persona-adaptive tutorial sequencer so that players may learn to adapt to different playstyles as the level context and difficulty changes. Tutorial generation is an exciting new paradigm, and the development of systems such as this one moves us closer toward fully automated video game tutorial generation.



\bibliographystyle{ACM-Reference-Format}
\bibliography{biblography}


\end{document}